\documentclass[conference]{IEEEtran}
\IEEEoverridecommandlockouts
\usepackage{cite}
\usepackage{amsmath,amssymb,amsfonts}
\usepackage{algorithmic}
\usepackage{graphicx}
\usepackage{textcomp}
\usepackage{xcolor}
\usepackage{multirow}
\def\BibTeX{{\rm B\kern-.05em{\sc i\kern-.025em b}\kern-.08em
    T\kern-.1667em\lower.7ex\hbox{E}\kern-.125emX}}
\begin{document}

\title{Revisiting Invariant Learning for Out-of-Domain Generalization on Multi-Site Mammogram Datasets\\
\thanks{Identify applicable funding agency here. If none, delete this.}
}

\author{Hung Q. Vo, Samira Zare, Son T. Ly, Lin Wang, Chika F. Ezeana, Xiaohui Yu, \\Kelvin K. Wong, Stephen T.C. Wong*, and Hien V. Nguyen*

\thanks{* Hien V. Nguyen and Stephen T.C. Wong are co-senior authors}
\thanks{Hung Q. Vo, Samira Zare, Son T. Ly, and Hien V. Nguyen are with the Department of Electrical and Computer Engineering, University of Houston, Houston, TX 77204 USA.}
\thanks{Lin Wang, Chika F. Ezeana, Xiaohui Yu, Kelvin K. Wong, and Stephen T.C. Wong are with Department of Systems Medicine and Biomedical Engineering, Houston Methodist Cancer Center, Houston, TX 77030 USA.}
}

\maketitle

\begin{abstract}
Achieving health equity in Artificial Intelligence (AI) requires diagnostic models that maintain reliability across diverse populations. However, breast cancer screening systems frequently suffer from domain overfitting, degrading significantly when deployed to varying demographics. While Invariant Learning algorithms aim to mitigate this by suppressing site-specific correlations, their efficacy in medical imaging remains underexplored. This study comprehensively evaluates domain generalization techniques for mammography.

We constructed a multi-source training environment aggregating datasets from the United States (CBIS-DDSM, EMBED), Portugal (INbreast, BCDR), and Cyprus (BMCD). To assess global generalizability, we evaluated performance on unseen cohorts from Egypt (CDD-CESM) and Sweden (CSAW-CC). We benchmarked Invariant Risk Minimization (IRM) and Variance Risk Extrapolation (VREx) against a rigorously optimized Empirical Risk Minimization (ERM) baseline. Contrary to expectations, standard ERM consistently outperformed specialized invariant mechanisms on out-of-domain testing. While VREx showed potential in stabilizing attention maps, invariant objectives proved unstable and prone to underfitting. We conclude that engineering equitable AI is currently best served by maximizing multi-national data diversity rather than relying on complex algorithmic invariance.
\end{abstract}


\section{Introduction}
Achieving health equity in Artificial Intelligence (AI) requires diagnostic models that maintain reliability across diverse patient populations and clinical environments. While deep learning has achieved expert-level performance in breast cancer screening \cite{yala2021toward, lotter2021robust}, these systems frequently suffer from domain overfitting—learning statistical shortcuts specific to a single hospital or demographic group. Consequently, performance often degrades significantly when models are deployed to under-represented populations or differing acquisition protocols, exacerbating existing healthcare disparities.

Recent empirical studies have highlighted the severity of this "generalization gap." For instance, Barros et al. \cite{barros2022virtual} observed sharp performance declines when models trained on US cohorts were evaluated on Israeli populations, attributing the failure to variations in health policies and demographics. Similarly, De et al. \cite{de2023impact} demonstrated that even commercial AI tools are fragile to changes in mammography equipment, often requiring local calibration to function correctly. These findings underscore a critical limitation: standard training methods (Empirical Risk Minimization) often latch onto "spurious correlations" (e.g., scanner artifacts or site-specific tissue density distributions) rather than learning the causal biology of cancer.

To mitigate these disparities, Invariant Learning algorithms, such as Invariant Risk Minimization (IRM) \cite{arjovsky2019invariant} and Variance Risk Extrapolation (VREx) \cite{krueger2021out}, have been proposed. Theoretically, these methods encourage models to ignore site-specific noise and focus on features that are stable across all environments. However, while applied to limited settings like lesion crops \cite{wang2022domain}, the practical efficacy of these invariant penalties in high-dimensional, whole-mammogram analysis remains underexplored.

In this paper, we conduct a comprehensive evaluation of domain generalization techniques to engineer more equitable screening tools. Unlike previous studies limited to single-site or two-continent comparisons, we construct a diverse multi-source training environment aggregating public datasets from the USA, Portugal, and Cyprus. We benchmark invariant strategies against a rigorously optimized standard baseline on unseen out-of-domain cohorts from Egypt and Sweden. Our goal is to determine whether complex algorithmic invariance is necessary for robustness, or if a data-centric approach—maximizing the diversity of training data—remains the most effective path toward generalized breast cancer prediction.

\section{Methods}
\subsection{Mammograms classification with tiled patches classifier initialization using CNNs}
In the domain of whole mammogram classification, we build upon prior research that has demonstrated the effectiveness of CNNs \cite{shen2019deep, sechopoulos2021artificial, sharma2021retrospective}. In line with the approach presented in \cite{shen2019deep}, our methodology involves initial training of the network using tiled patches, encompassing both lesion patches and background crops. Subsequently, the pre-trained network serves as the starting point for initializing the whole mammogram network. In this study, we employ a standard residual network (ResNet) \cite{he2016deep}, and a more modernized ConvNet (ConvNeXt) \cite{liu2022convnet}, which has demonstrated competitive accuracy and scalability compared to the Vision Transformer \cite{dosovitskiy2020image}.

\begin{table*}[t]
\centering
\setlength{\tabcolsep}{6pt}
\caption{Comparative evaluation of Standard ERM against domain-invariant strategies (IRM, VREx) for whole-mammogram cancer classification. Experiments utilize ResNet34 and ConvNeXt-tiny backbones on both in-domain (CBIS-DDSM, EMBED) and out-of-domain (CDD-CESM, CSAW-CC) test sets. Results indicate that Standard ERM remains highly competitive, particularly in OOD scenarios. (*: ERM approximated by setting the IRM penalty term to zero, which is usually used as a baseline in invariant learning papers).}
\label{table:experiments}
\begin{tabular}{|c|l|ccc|ccc|ccc|ccc|}
\hline
\multirow{2}{*}{\textbf{Model}} & \multicolumn{1}{c|}{\multirow{2}{*}{\textbf{Method}}} & \multicolumn{6}{c|}{\textbf{In-Domain}} & \multicolumn{6}{c|}{\textbf{Out-of-Domain}} \\ \cline{3-14} 
 & & \multicolumn{3}{c|}{CBIS-DDSM} & \multicolumn{3}{c|}{EMBED} & \multicolumn{3}{c|}{CDD-CESM} & \multicolumn{3}{c|}{CSAW-CC} \\ \cline{3-14} 
 & & Acc & AP & AUC & Acc & AP & AUC & Acc & AP & AUC & Acc & AP & AUC \\ \hline
 
\multirow{4}{*}{ResNet34}      
& ERM \cite{vapnik1991principles} & 0.648 & 0.729 & 0.785 & 0.703 & 0.437 & 0.524 & 0.733 & \textbf{0.651} & \textbf{0.740} & 0.984 & \textbf{0.081} & \textbf{0.664} \\
& ERM* & 0.566 & \textbf{0.745} & 0.788 & 0.738 & \textbf{0.559} & 0.630 & 0.692 & 0.508 & 0.652 & 0.979 & 0.032 & 0.549 \\
& IRM \cite{arjovsky2019invariant} & 0.602 & 0.388 & 0.460 & 0.609 & 0.266 & 0.391 & 0.673 & 0.289 & 0.436 & 0.990 & 0.010 & 0.495 \\
& VREx \cite{krueger2021out} & 0.570 & 0.741 & \textbf{0.797} & 0.641 & 0.521 & \textbf{0.641} & 0.644 & 0.545 & 0.672 & 0.914 & 0.059 & 0.628 \\ \hline

\multirow{4}{*}{\shortstack{ConvNeXt\\-tiny}} 
& ERM \cite{vapnik1991principles} & 0.733 & \textbf{0.742} & \textbf{0.797} & 0.814 & \textbf{0.719} & 0.746 & 0.794 & \textbf{0.731} & \textbf{0.832} & 0.979 & \textbf{0.108} & \textbf{0.727} \\
& ERM* & 0.712 & 0.719 & 0.759 & 0.795 & 0.676 & 0.705 & 0.759 & 0.697 & 0.798 & 0.980 & 0.070 & 0.669 \\
& IRM \cite{arjovsky2019invariant} & 0.401 & 0.527 & 0.590 & 0.326 & 0.513 & 0.604 & 0.618 & 0.577 & 0.709 & 0.989 & 0.019 & 0.580 \\
& VREx \cite{krueger2021out} & 0.638 & 0.731 & 0.775 & 0.762 & 0.671 & \textbf{0.748} & 0.729 & 0.679 & 0.785 & 0.953 & 0.064 & 0.596 \\ \hline
\end{tabular}
\end{table*}

\subsection{Invariant Learning algorithms}
\subsubsection{Invariant Risk Minimization (IRM)}
Invariant Risk Minimization (IRM) \cite{arjovsky2019invariant} is a learning approach that focuses on estimating consistent correlations across diverse training distributions. By learning a data representation that ensures an optimal classifier performs well across all training distributions, IRM enables effective generalization to out-of-distribution scenarios. This capability is particularly valuable in real-world applications such as medical diagnosis, where the data distribution can vary significantly across different hospitals and institutions. The invariances captured by IRM are associated with the underlying causal structures that govern the data, enhancing its ability to uncover meaningful patterns and relationships.

Let $e \in \mathcal{E}$ be an environment with its own data distribution $\mathcal{X}, \mathcal{Y}$. Denote $\mathcal{E}_{\mathrm{tr}} \subset \mathcal{E}$, $\mathcal{E}_{\mathrm{tr}}$ is an empirical set of environments or a set of environments that can be accessed. Let $\Phi: \mathcal{X} \rightarrow \mathcal{H}$ be a projection function of data into the latent space. $R^e(\Phi)$ will be a risk of environment $e$ under $\Phi$. The practical IRM objective is:
\begin{equation}
    \Phi^*=\underset{\Phi: \mathcal{X} \rightarrow \mathbb{R}^d}{\arg \min } \sum_{e \in \mathcal{E}_{\mathrm{tr}}} R^e(\Phi)+\lambda\left(\left.\nabla_w R^e(w \cdot \Phi)\right|_{w=1}\right)^2
\end{equation}
Here, $w$ is a simplified version of the predictor $g: \mathcal{H} \rightarrow \mathcal{Y}$ which is assumed to be optimal over all environments. To streamline the learning process, the value of $w$ is fixed at 1, allowing the network to concentrate on learning the representations captured by $\Phi$.

\subsubsection{Variance Risk Extrapolation (VREx)}
Risk Extrapolation (REx) \cite{krueger2021out} demonstrates superior performance compared to Invariant Risk Minimization (IRM) when faced with scenarios where both causally induced distributional shifts and covariate shifts occur simultaneously. REx achieves this by effectively balancing its resilience to these types of shifts. In contrast, IRM assumes that the learned invariances from the training data alone are adequate for generalization to new domains, which may not always hold true. While both REx and IRM are capable of performing causal identification, REx demonstrates greater robustness in handling covariate shifts (changes in the input distribution). Covariate shift poses challenges when models are misspecified, when training data is limited, or when it fails to cover relevant areas of the test distribution.
 
REx is a robust optimization technique that operates within a perturbation set of extrapolated domains. Another variant, named VREx, introduces a penalty term that targets the variance of training risks. It has been shown that variants of REx can recover the causal mechanisms of the targets, while also providing some robustness to changes in the input distribution. VREx shares a similar objective to IRM but replaces the penalty term with the variance of training risks across different environments:
\begin{equation}
    \Phi^*=\underset{\Phi: \mathcal{X} \rightarrow \mathbb{R}^d}{\arg \min }[\lambda\left.Var( \{R^e(\Phi)\}_{e \in \mathcal{E}_{\mathrm{tr}}}\right)+\sum_{e \in \mathcal{E}_{\mathrm{tr}}} R^e(\Phi)]
\end{equation}

\subsubsection{Invariant Learning on multi-site datasets}
In our study, we adopt the perspective that each dataset obtained from a particular hospital or institution represents a distinct environment. The training process involves straightforward computation of risks for each dataset, followed by the calculation of either the IRM or VREx objective. To ensure the validity of our approach, we ensure that the datasets used for training and testing originate from different sites or institutions. This approach enables us to evaluate the performance of our models across diverse environments and establish their robustness in handling variations between different data sources \cite{gulrajani2020search}.

\section{Experiments}
\subsection{Datasets}
To ensure reproducibility and facilitate future benchmarking, this study exclusively utilizes publicly available datasets. We aggregated data from seven diverse sources: CBIS-DDSM (USA) \cite{lee2017curated}, EMBED (USA) \cite{jeong2023emory}, INbreast (Portugal) \cite{moreira2012inbreast}, BMCD (Republic of Cyprus) \cite{loizidou2021digital}, BCDR (Portugal) \cite{moura2013benchmarking}, CDD-CESM (Egypt) \cite{khaled2022categorized}, and CSAW-CC (Sweden) \cite{Strand2022CSAW-CC}.

\textbf{Experimental Protocol:}
We designed our evaluation to assess both in-domain and out-of-domain (OOD) performance.
\begin{itemize}
    \item \textbf{Training Domain:} The primary training corpus consists of CBIS-DDSM, EMBED, INbreast, BMCD, and BCDR.
    \item \textbf{Out-of-Domain Testing:} To evaluate generalization capability, we reserved the CDD-CESM and CSAW-CC datasets entirely for testing, as they originate from geographic regions not represented in the training phase.
    \item \textbf{In-Domain Testing:} To monitor standard performance, we utilized the official test set of CBIS-DDSM and a held-out 25\% subset of the EMBED dataset.
\end{itemize}
Strict patient-level separation was enforced across all splits to prevent data leakage.

\subsection{Experimental Settings}
\textbf{Architectures:} We benchmark two Convolutional Neural Network (CNN) backbones: ResNet34 and ConvNeXt-tiny. While ResNet34 serves as a standard baseline, ConvNeXt-tiny was selected for its ability to achieve competitive accuracy and scalability comparable to Vision Transformers (ViTs) while maintaining the inductive biases of CNNs \cite{liu2022convnet}.

\textbf{Training Strategy:} To address the high resolution and data scarcity inherent in mammography, we adopt the two-stage transfer learning strategy proposed by Shen et al. \cite{shen2019deep}.
\begin{enumerate}
    \item \textbf{Patch Pre-training:} We first train a classifier on ROI-annotated patches (lesions and background) extracted from CBIS-DDSM \cite{lee2017curated}, EMBED \cite{jeong2023emory}, INbreast \cite{moreira2012inbreast}, and BCDR \cite{moura2013benchmarking}. This initializes the network with discriminative local features.
    \item \textbf{Whole-Mammogram Fine-tuning:} The pre-trained weights are then transferred to the whole-mammogram classifier, which is fine-tuned on the full images.
\end{enumerate}

\textbf{Implementation Details:} All models are optimized using Adam with a learning rate of $1\mathrm{e}{-3}$. Due to computational constraints, input mammograms are resized to $576 \times 448$. We acknowledge that while higher resolutions (e.g., $1152 \times 896$ \cite{shen2019deep} or larger \cite{wu2019deep}) often yield performance gains, our resolution choice balances resource efficiency with representational capacity. The quantitative results of these experiments are summarized in Table \ref{table:experiments}.

\subsection{Quantitative Analysis}
Table \ref{table:experiments} details the performance of ResNet34 and ConvNeXt-tiny across in-domain and out-of-domain (OOD) settings. We compared four training paradigms: Standard ERM \cite{vapnik1991principles} (fully optimized), Approximated ERM* (IRM with $\lambda=0$), IRM \cite{arjovsky2019invariant}, and VREx \cite{krueger2021out}.

\textbf{Robustness of Standard ERM:} A critical finding of our study is the resilience of Standard ERM when trained on diverse data. On the unseen out-of-domain datasets (CDD-CESM and CSAW-CC), Standard ERM consistently achieved the highest generalization performance. With the ConvNeXt-tiny architecture, Standard ERM reached an AUROC of 0.832 on CDD-CESM and 0.727 on CSAW-CC, outperforming both the approximated baseline (ERM*) and the invariant methods. This highlights the importance of using a fully optimized Standard ERM as a rigorous benchmark when evaluating the added value of domain generalization techniques.

\textbf{Instability of Invariant Constraints:} Contrary to theoretical expectations, IRM consistently yielded the lowest performance, often trailing significantly behind the baselines. An analysis of training dynamics suggests severe underfitting; the complex penalty landscape of IRM appears difficult to traverse in high-dimensional mammography, leading to optimization instability. VREx showed better stability than IRM, occasionally matching ERM on in-domain tasks (e.g., ResNet34 on CBIS-DDSM), but it failed to provide a generalization benefit on the challenging OOD cohorts.

\textbf{Architecture Impact:} ConvNeXt-tiny consistently outperformed ResNet34 across all metrics and datasets. This performance gap was particularly pronounced on the EMBED test set, confirming that modernizing the CNN backbone provides a more tangible reliability improvement than altering the loss function.

\subsection{Qualitative Analysis}
\begin{figure}[!t]
   \centering
\resizebox{\linewidth}{!}{
\setlength{\tabcolsep}{7pt}
\begin{tabular}{p{0.01cm}cc}

\rotatebox[origin=l]{90}{\small \textbf{CDD}} &

\shortstack{\small \textbf{ERM} \\ \includegraphics[width=0.5\linewidth,scale=0.5]{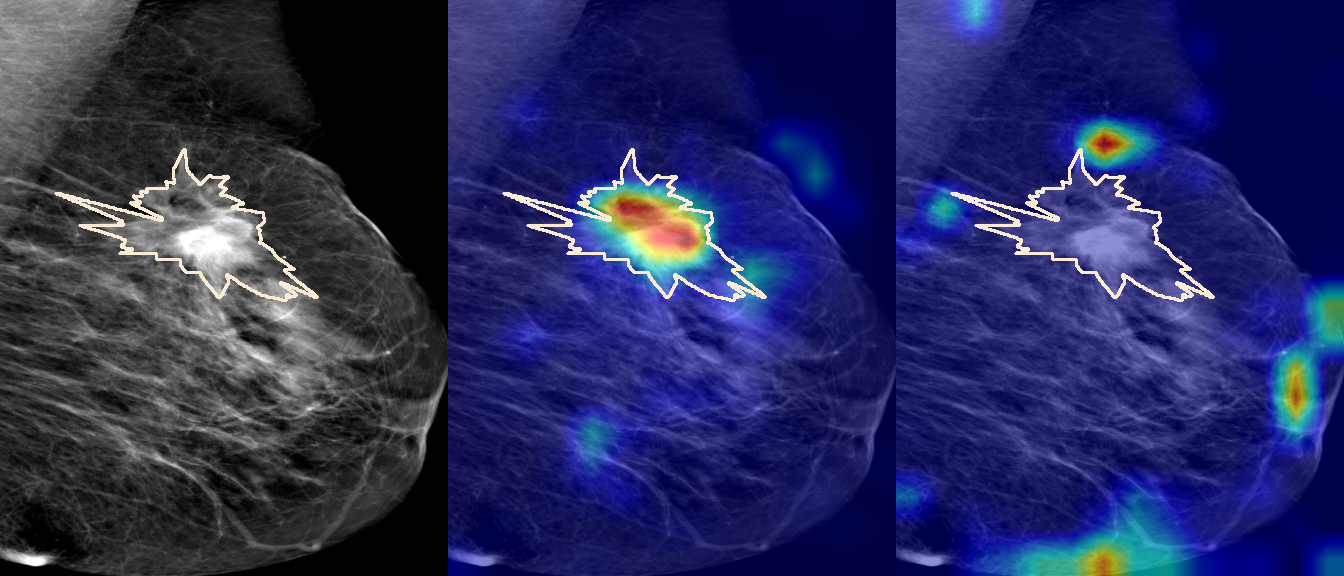}}&
\shortstack{\small \textbf{VREx} \\ \includegraphics[width=0.5\linewidth,scale=0.5]{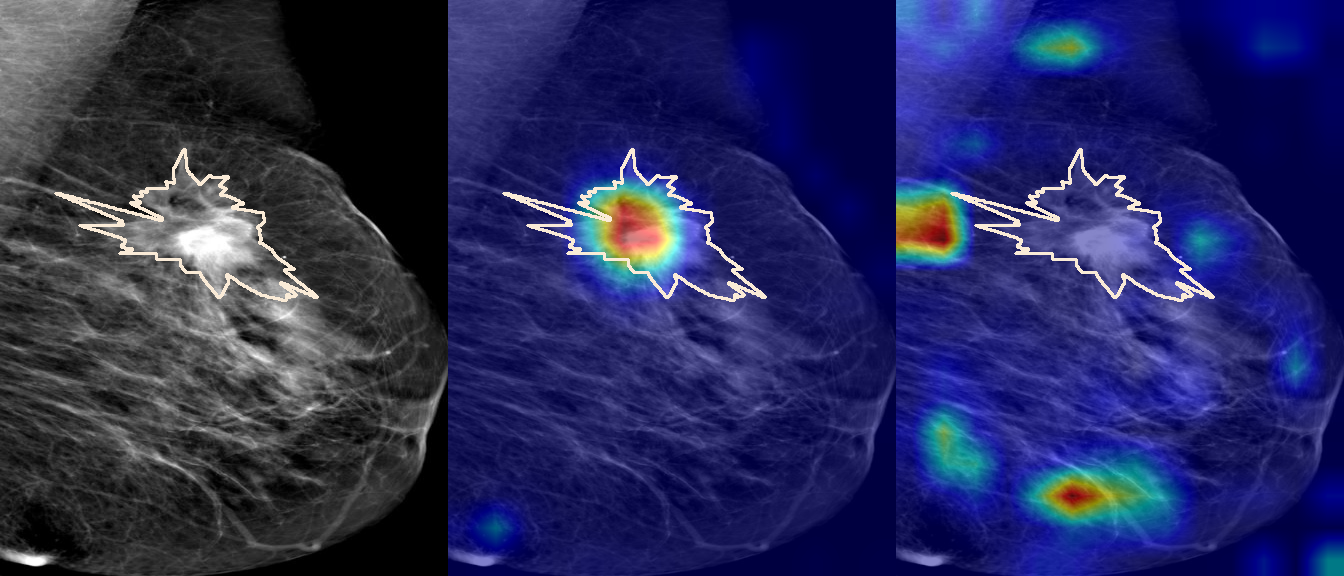}} \\ [1pt]

\rotatebox[origin=l]{90}{\small \textbf{CSAW}} &
\shortstack{\includegraphics[width=0.5\linewidth,scale=0.5]{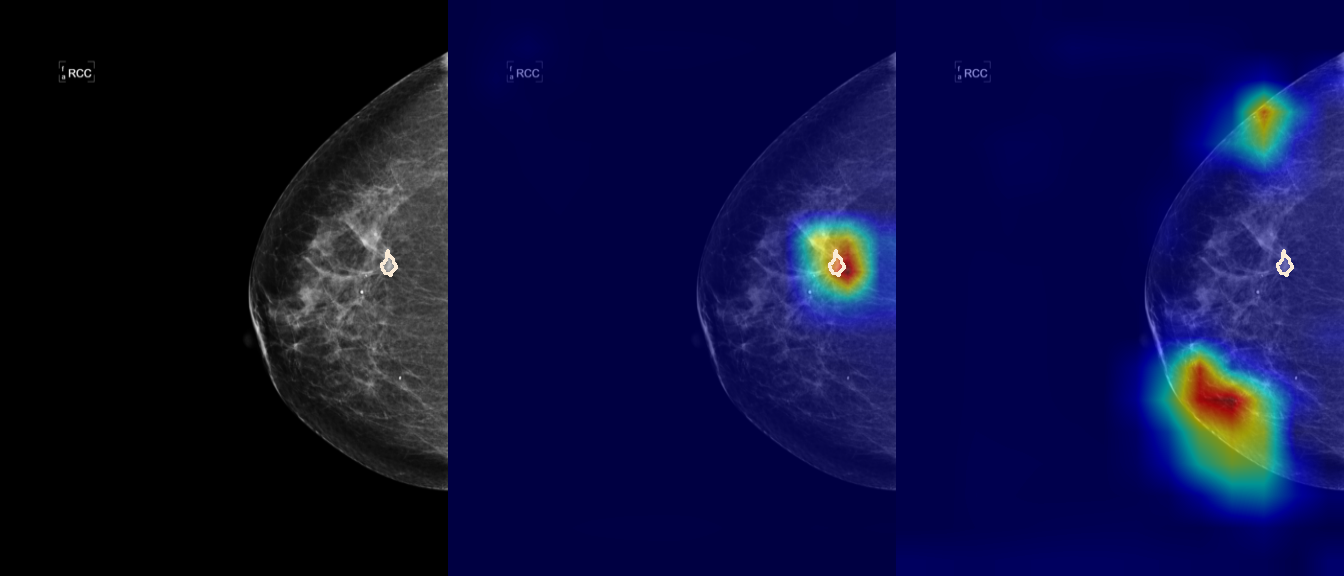}}&
\shortstack{\includegraphics[width=0.5\linewidth,scale=0.5]{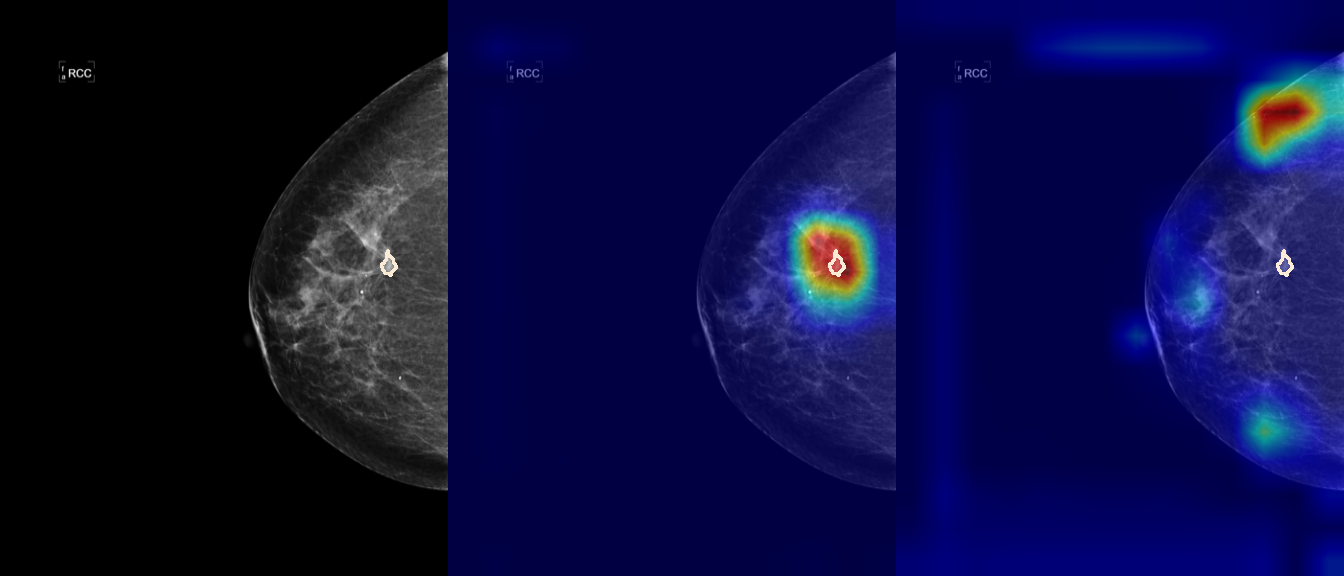}} \\ [1pt]

\rotatebox[origin=l]{90}{\small \textbf{CSAW}} &
\shortstack{\includegraphics[width=0.5\linewidth,scale=0.5]{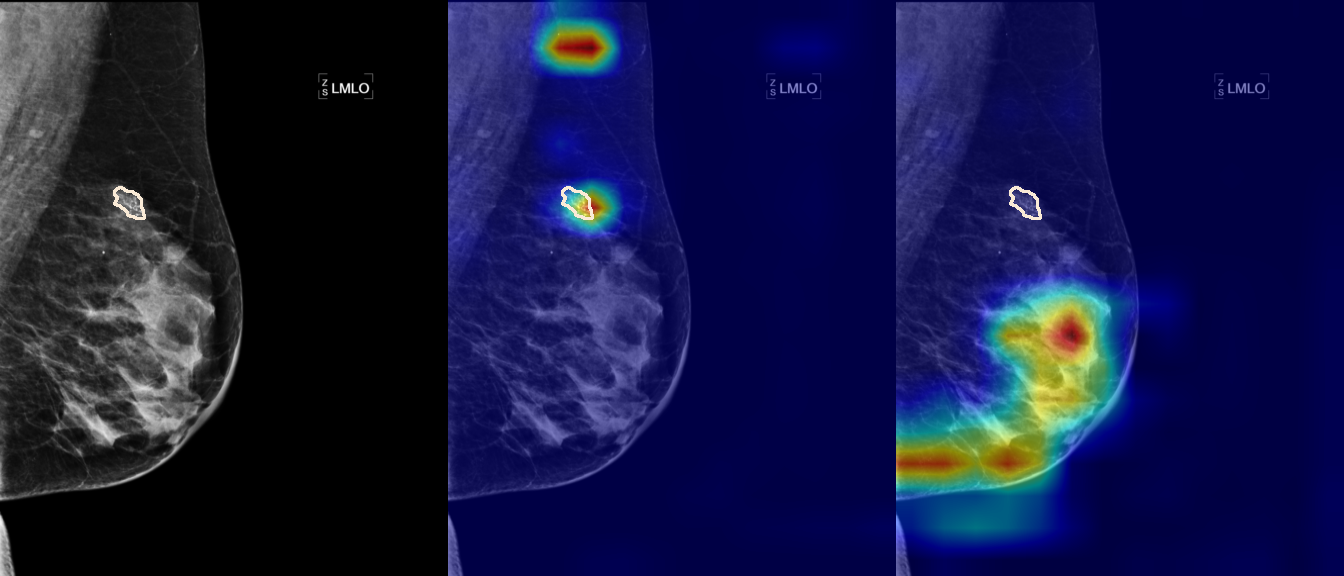}}&
\shortstack{\includegraphics[width=0.5\linewidth,scale=0.5]{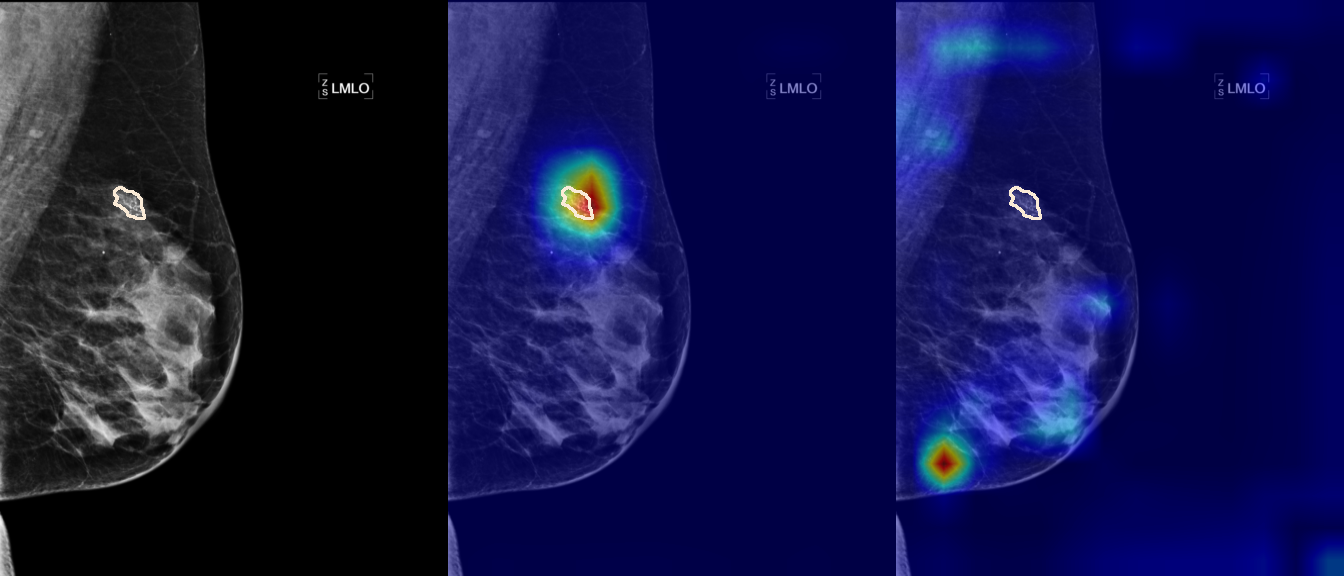}} \\ [1pt]

\rotatebox[origin=l]{90}{\small \textbf{EMBED}} &
\shortstack{\includegraphics[width=0.5\linewidth,scale=0.5]{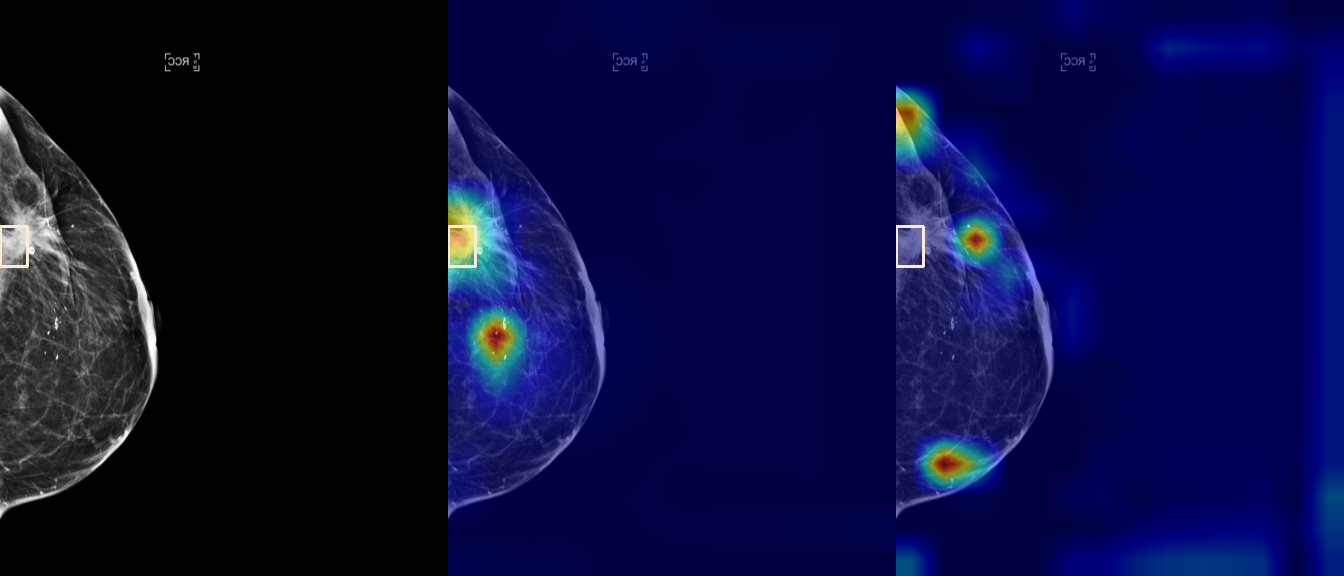}}&
\shortstack{\includegraphics[width=0.5\linewidth,scale=0.5]{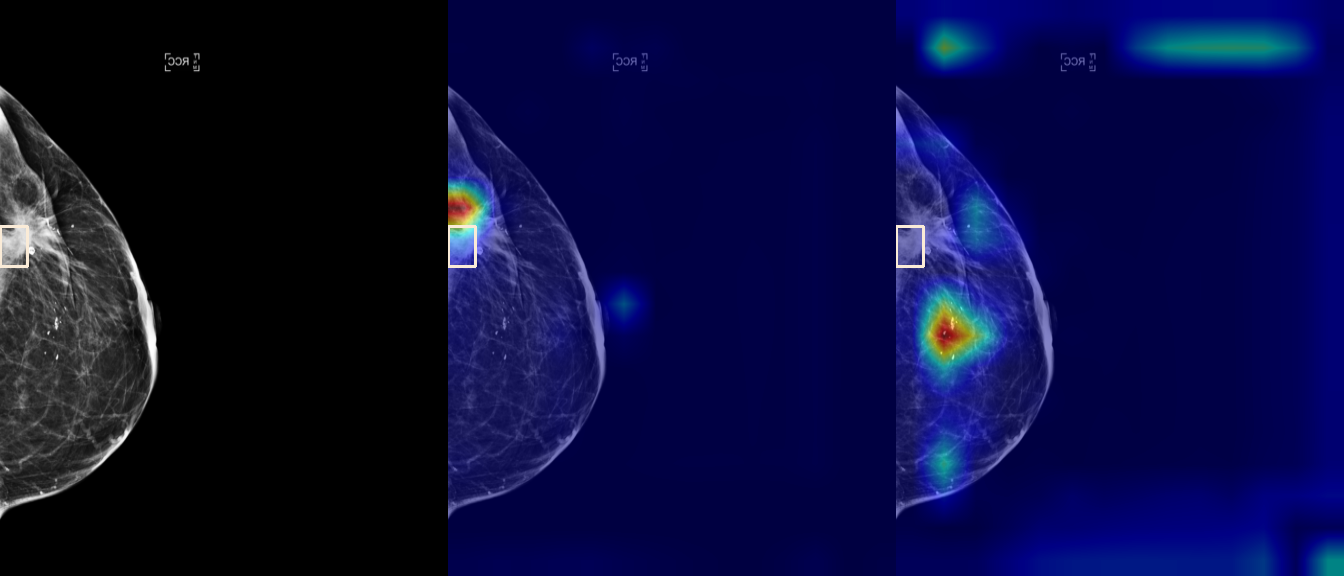}} \\ [1pt]

\rotatebox[origin=l]{90}{\small \textbf{CSAW}} &
\shortstack{\includegraphics[width=0.5\linewidth,scale=0.5]{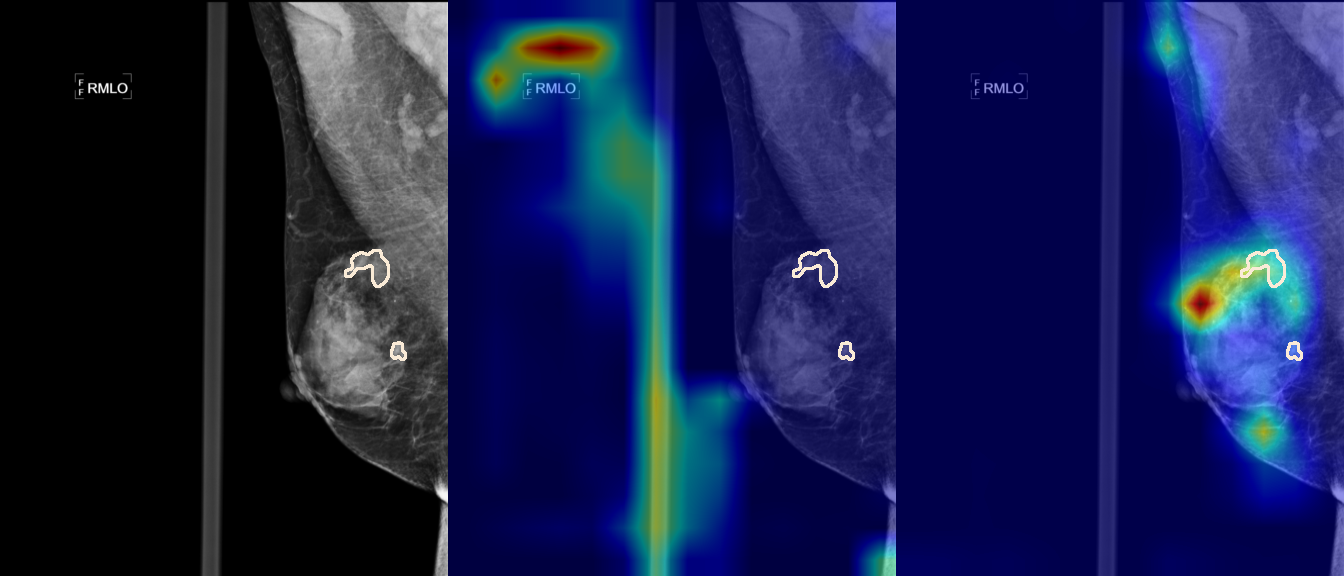}}&
\shortstack{\includegraphics[width=0.5\linewidth,scale=0.5]{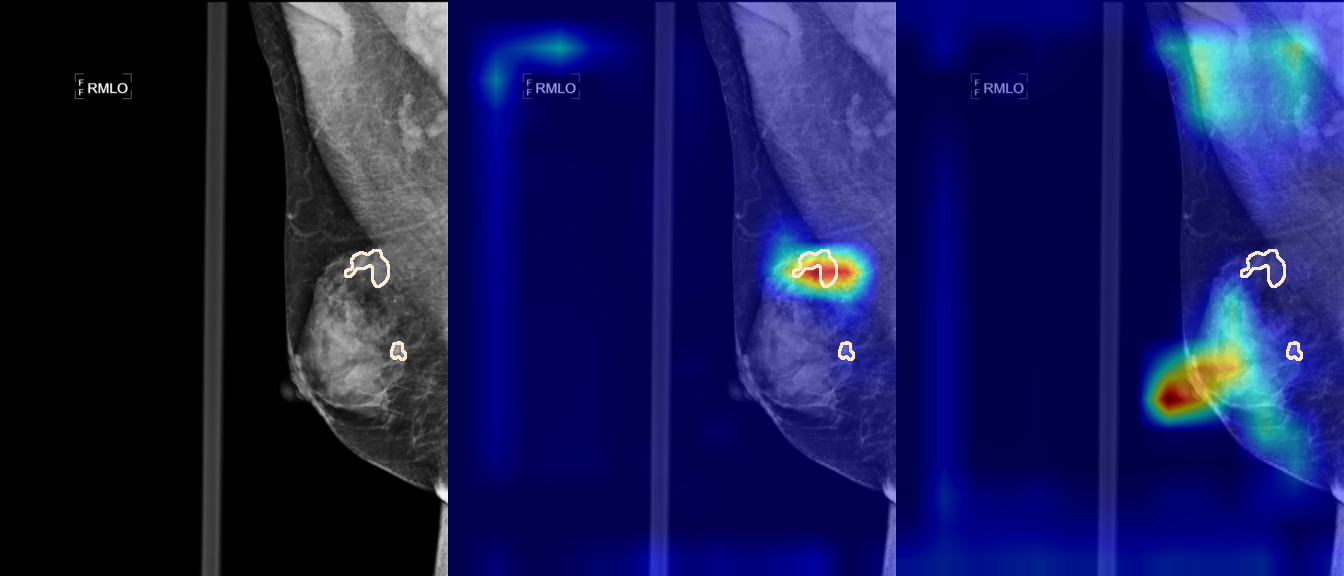}} \\ [1pt]

\rotatebox[origin=l]{90}{\small \textbf{CSAW}} &
\shortstack{\includegraphics[width=0.5\linewidth,scale=0.5]{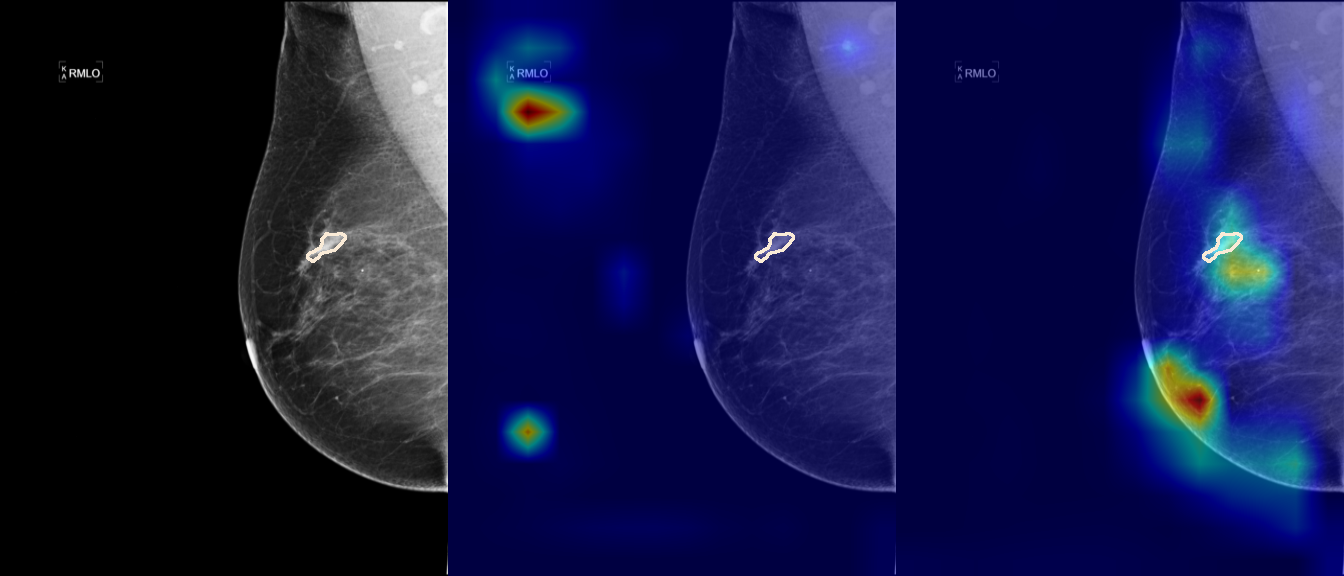}}&
\shortstack{\includegraphics[width=0.5\linewidth,scale=0.5]{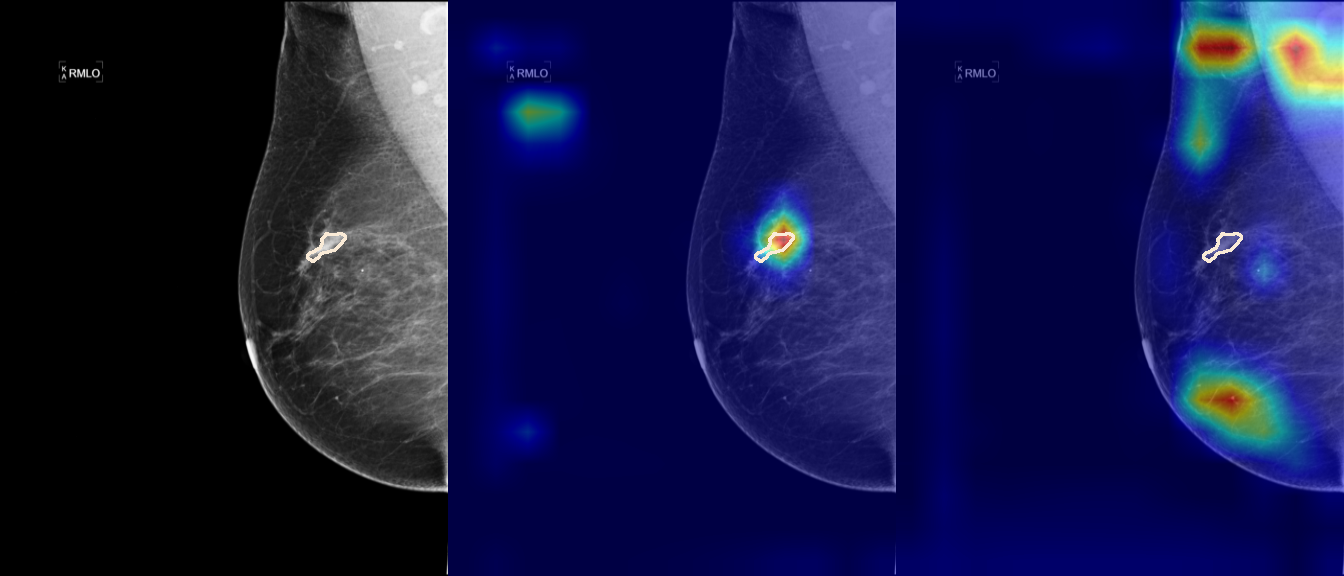}} \\ [1pt]

\end{tabular}
}
    \caption{Qualitative comparison of ResNet34d feature visualizations. Rows correspond to different datasets, while columns represent the training methods. Each entry displays a triplet: the original mammogram (left), the activation map for the Malignant class (center), and the activation map for the Benign class (right).}
    \label{fig:1_resnet34d_good_cases} 
\end{figure}

\begin{figure}[!t]
   \centering
\resizebox{\linewidth}{!}{
\setlength{\tabcolsep}{7pt}
\begin{tabular}{p{0.01cm}cc}

\rotatebox[origin=l]{90}{\small \textbf{CDD}} &
\shortstack{\small \textbf{ERM} \\ \includegraphics[width=0.5\linewidth,scale=0.5]{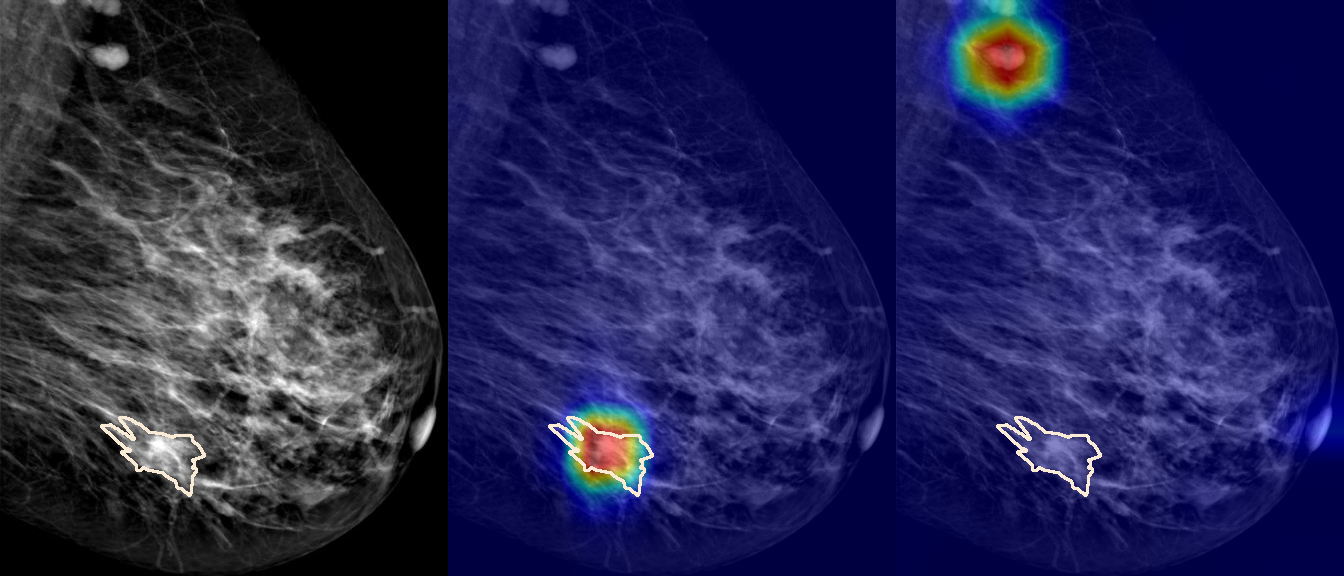}}&
\shortstack{\small \textbf{VREx} \\ \includegraphics[width=0.5\linewidth,scale=0.5]{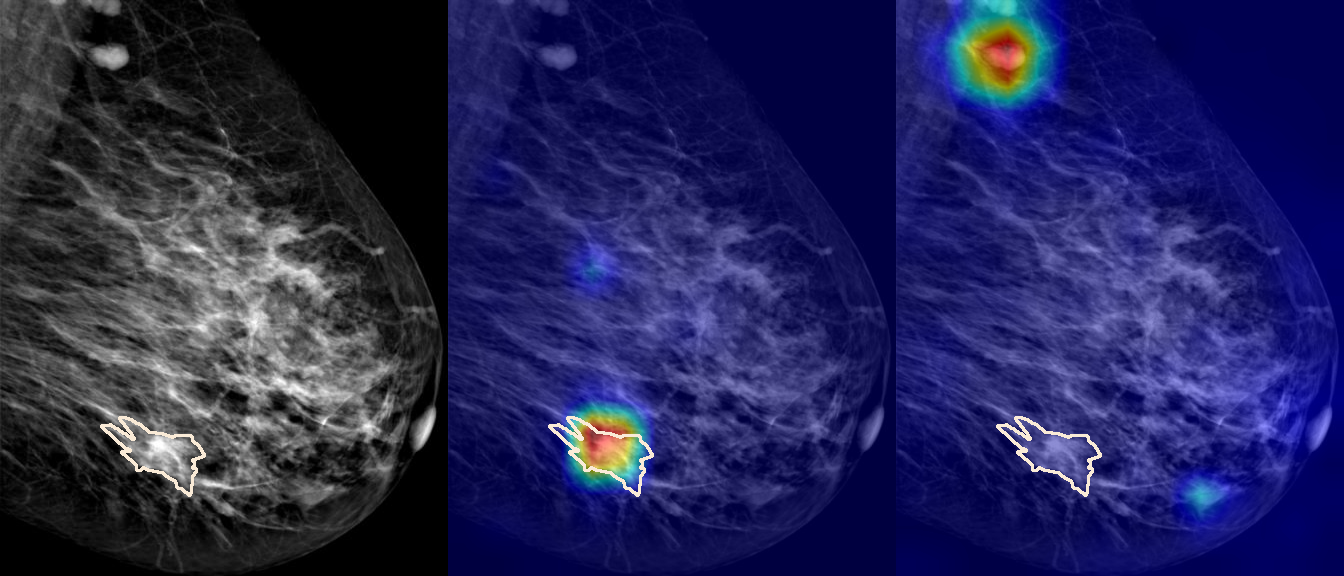}} \\ [1pt]

\rotatebox[origin=l]{90}{\small \textbf{CSAW}} &
\shortstack{\includegraphics[width=0.5\linewidth,scale=0.5]{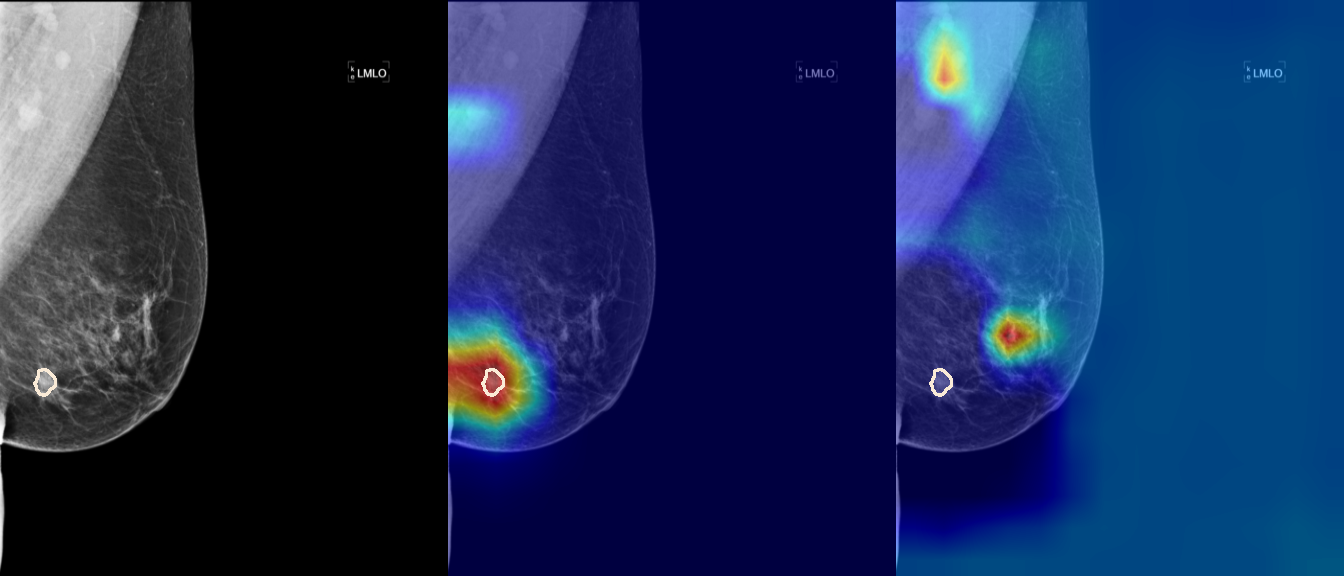}}&
\shortstack{\includegraphics[width=0.5\linewidth,scale=0.5]{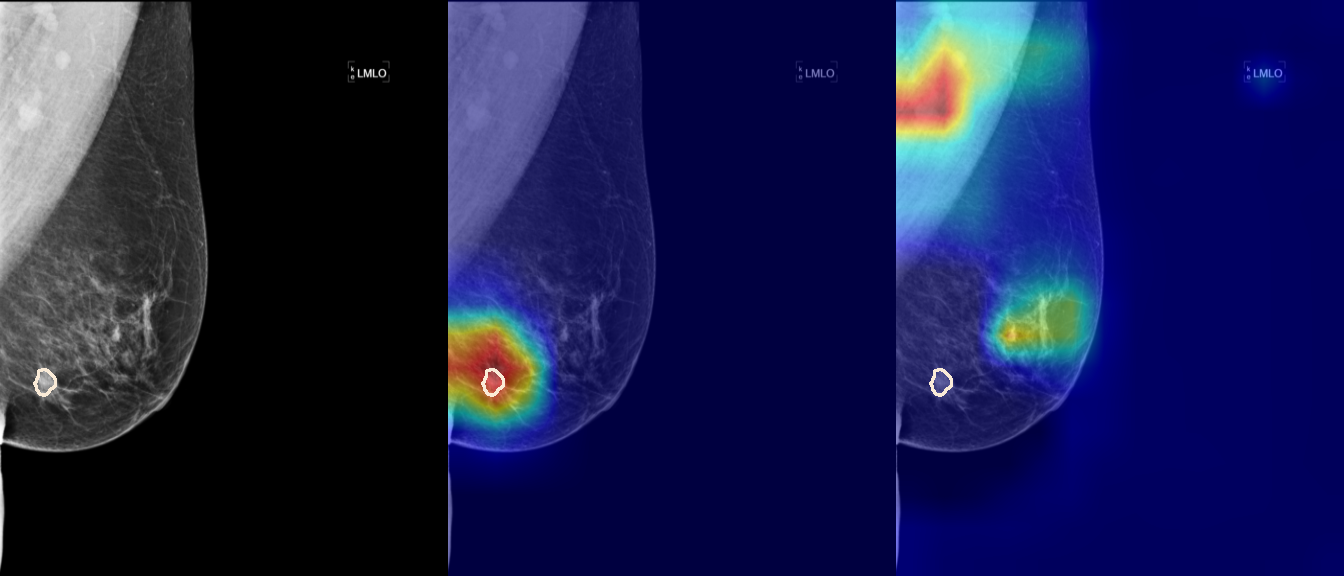}} \\ [1pt]

\rotatebox[origin=l]{90}{\small \textbf{CSAW}} &
\shortstack{\includegraphics[width=0.5\linewidth,scale=0.5]{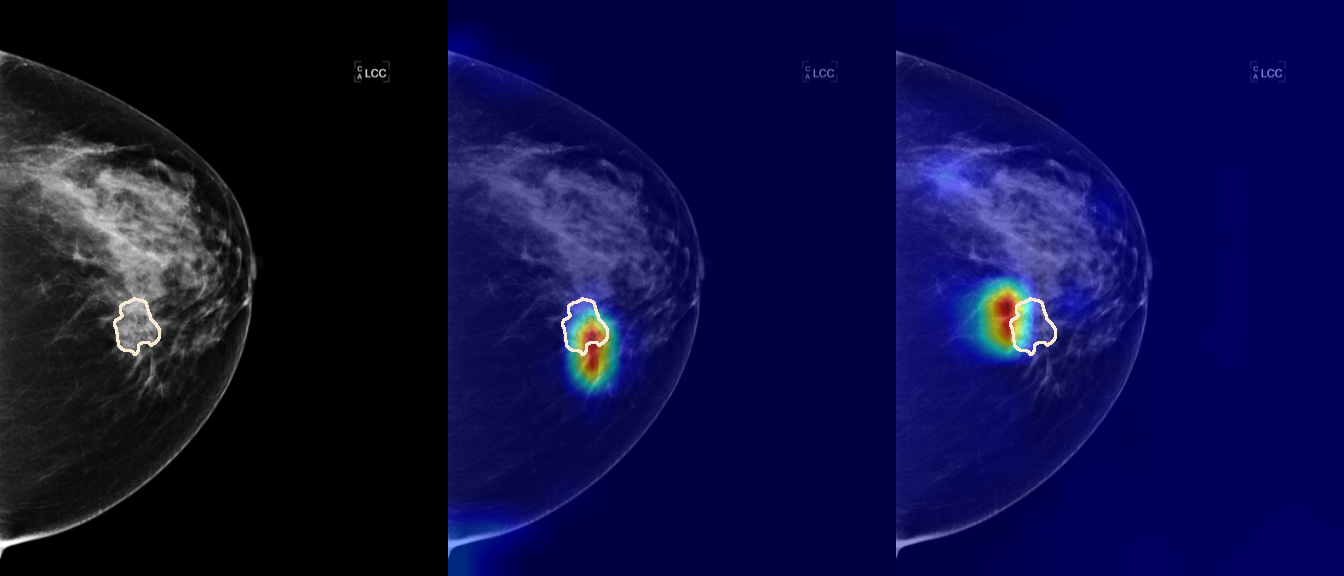}}&
\shortstack{\includegraphics[width=0.5\linewidth,scale=0.5]{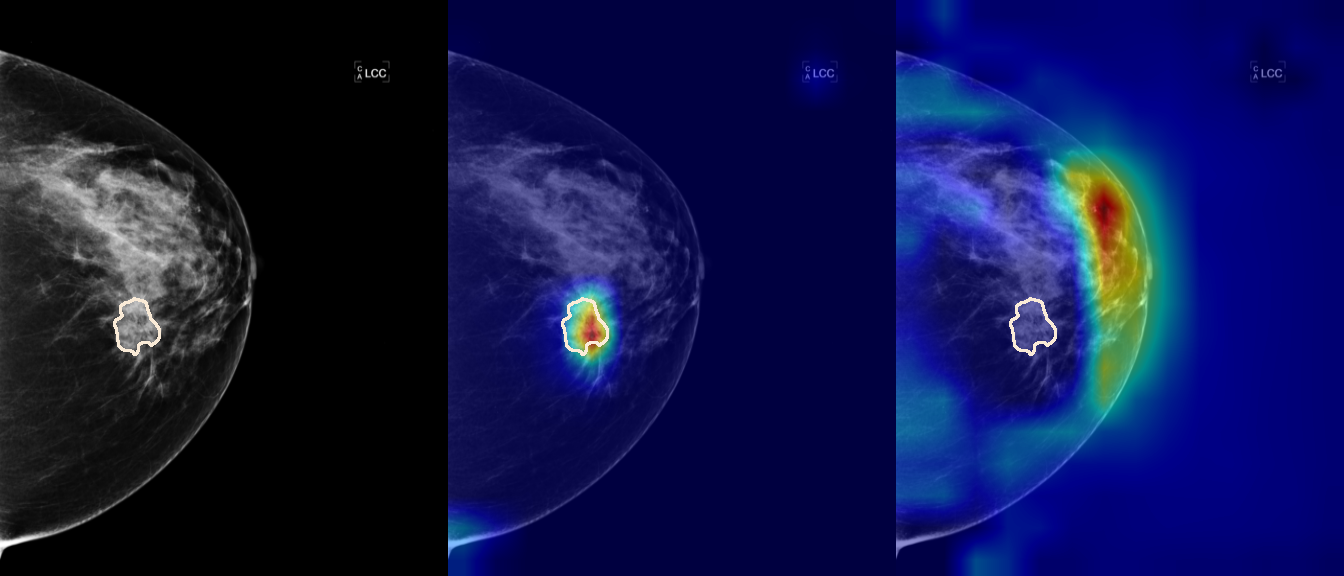}} \\ [1pt]

\rotatebox[origin=l]{90}{\small \textbf{EMBED}} &
\shortstack{\includegraphics[width=0.5\linewidth,scale=0.5]{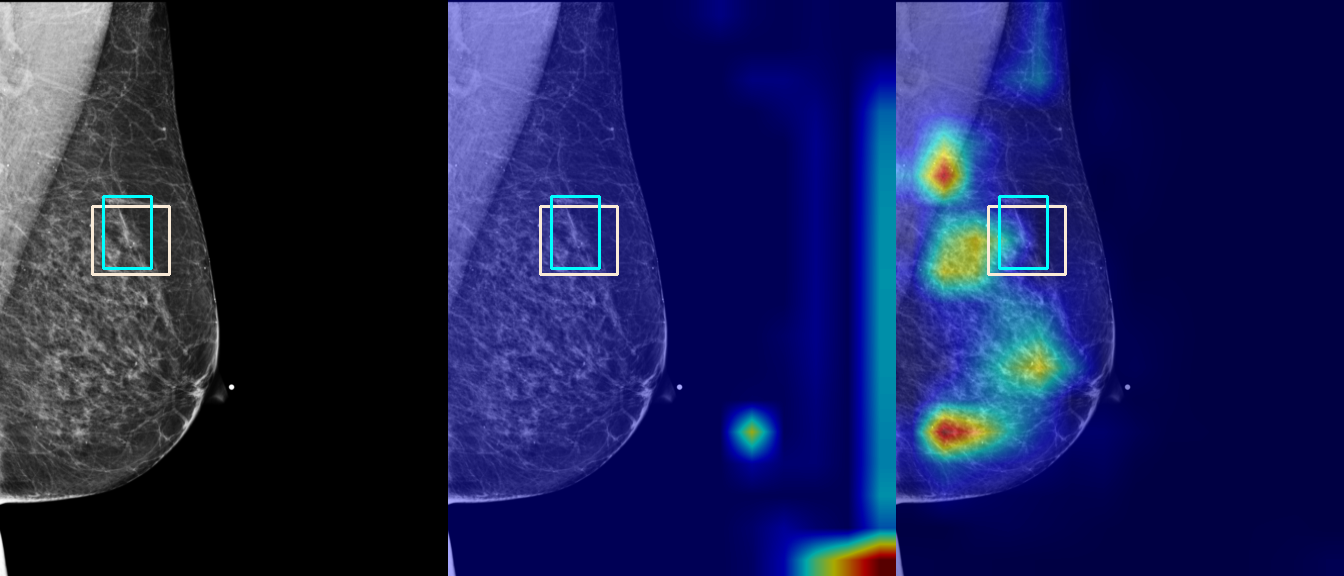}}&
\shortstack{\includegraphics[width=0.5\linewidth,scale=0.5]{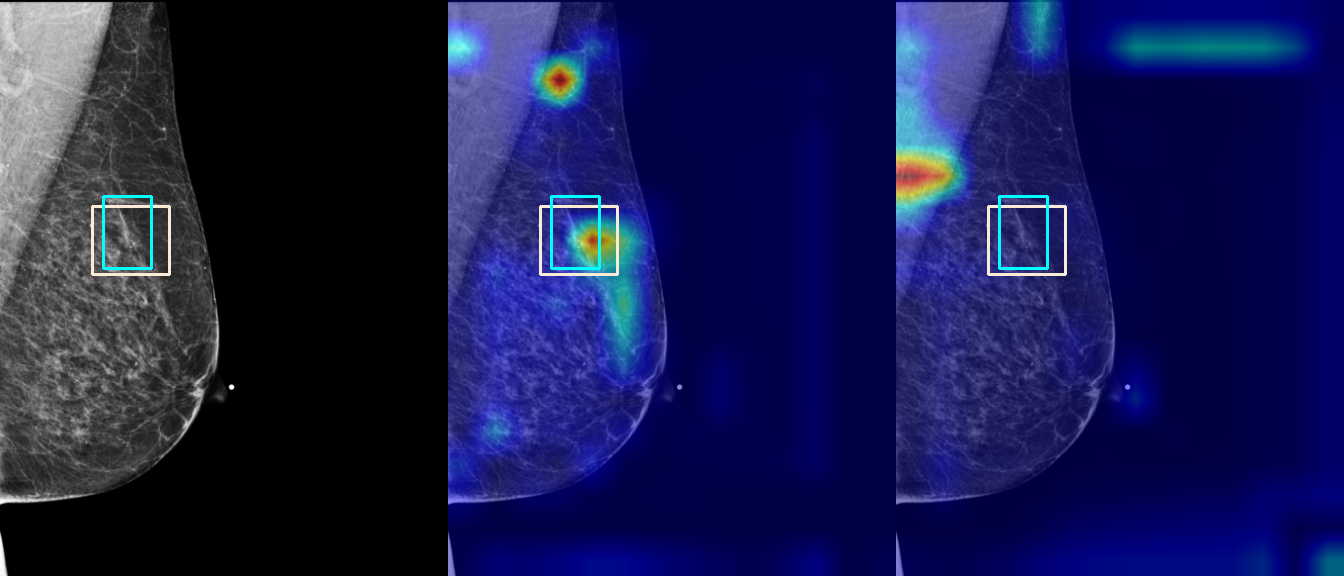}} \\ [1pt]

\rotatebox[origin=l]{90}{\small \textbf{CSAW}} &
\shortstack{\includegraphics[width=0.5\linewidth,scale=0.5]{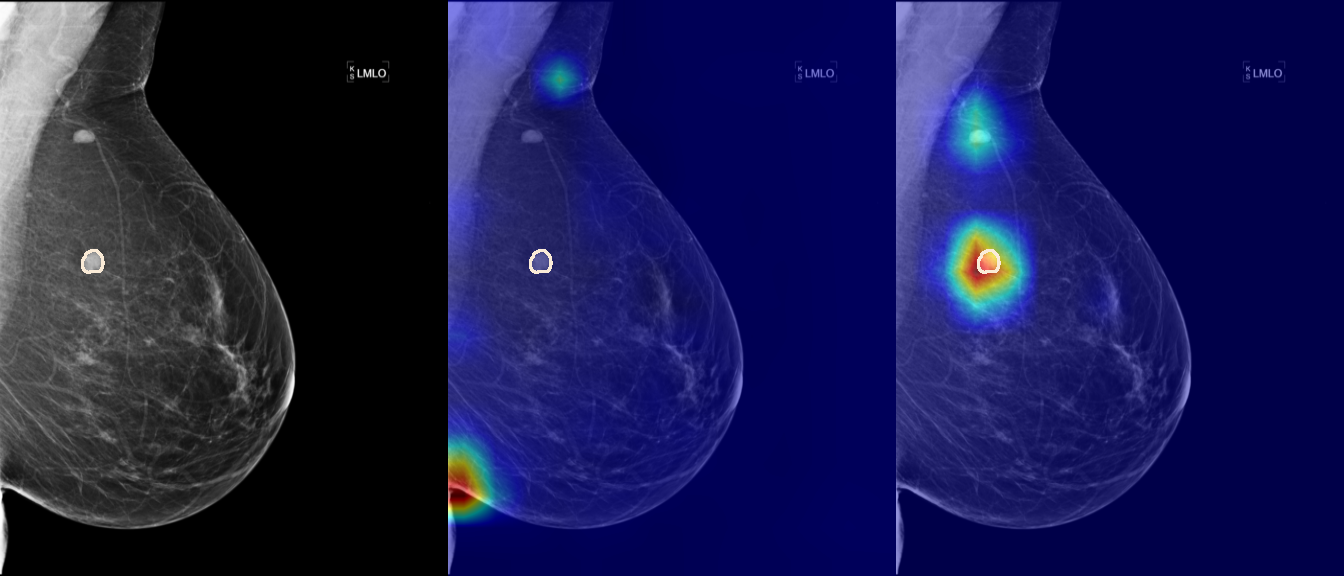}}&
\shortstack{\includegraphics[width=0.5\linewidth,scale=0.5]{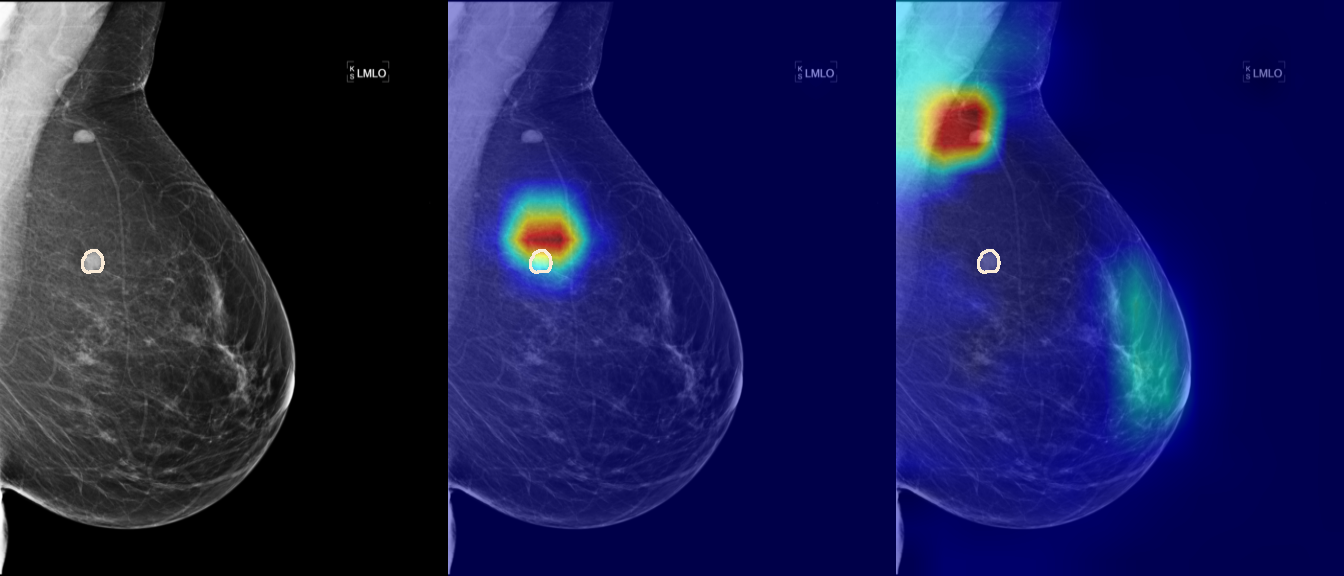}} \\ [1pt]

\rotatebox[origin=l]{90}{\small \textbf{CSAW}} &
\shortstack{\includegraphics[width=0.5\linewidth,scale=0.5]{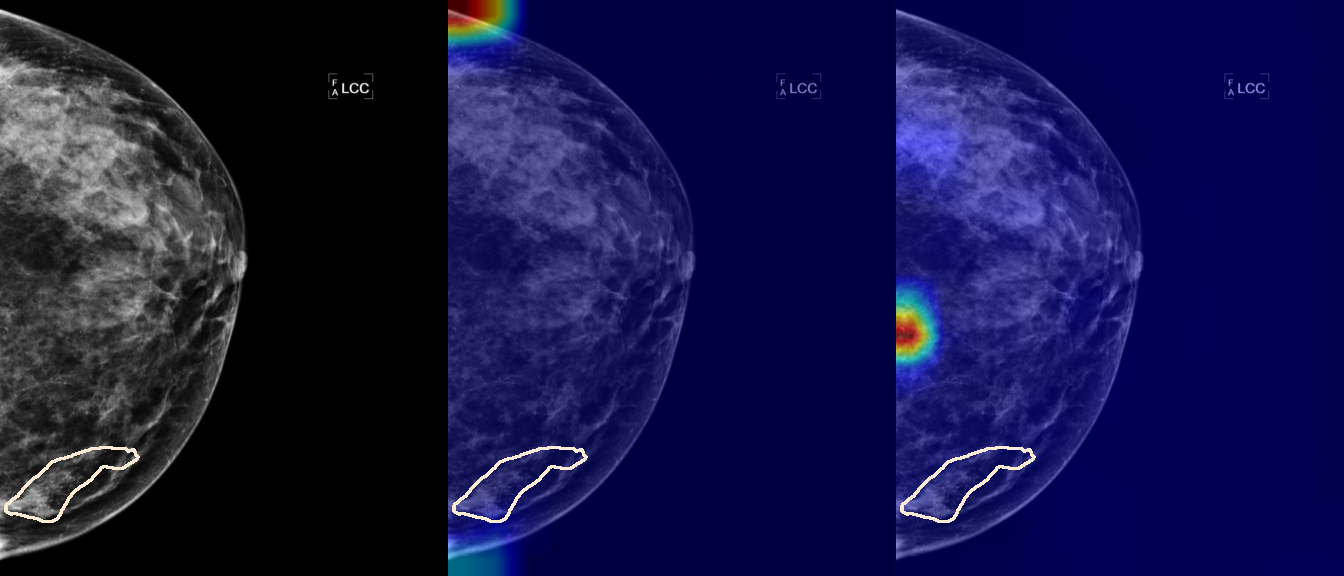}}&
\shortstack{\includegraphics[width=0.5\linewidth,scale=0.5]{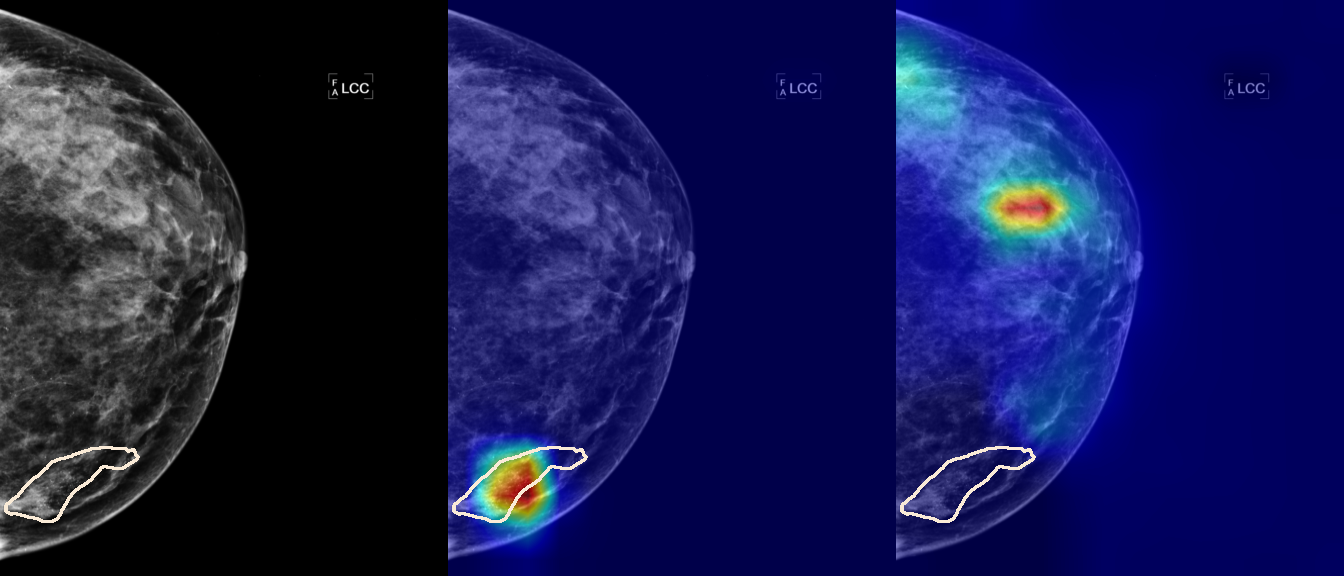}} \\ [1pt]

\end{tabular}
}
    \caption{Qualitative comparison of ConvNeXt-tiny feature visualizations. Rows correspond to different datasets, while columns represent the training methods. Each entry displays a triplet: the original mammogram (left), the activation map for the Malignant class (center), and the activation map for the Benign class (right).}
    \label{fig:3_convnext_tiny_good_cases} 
\end{figure}

To interpret the decision-making process behind the quantitative results, we visualized Class Activation Maps (CAM) for the top-performing methods, ERM and VREx. Figures \ref{fig:1_resnet34d_good_cases} through \ref{fig:3_convnext_tiny_good_cases} illustrate discriminative regions identified by the models within the full mammograms.

\textbf{Localization of Clinical Features:}
A key concern in medical AI is whether high performance stems from genuine pathology detection or spurious background correlations. Our visualizations suggest that Standard ERM achieves its superior quantitative performance by predominantly prioritizing clinical cues. As shown in the top rows of the Figure \ref{fig:1_resnet34d_good_cases} and Figure \ref{fig:3_convnext_tiny_good_cases}, the ERM-trained models consistently focus attention on the specific lesion regions within the breast tissue. This demonstrates that despite lacking explicit invariance constraints, ERM is capable of learning robust, medically relevant features when trained on diverse multi-source data.

\textbf{Attention Drift in Both Methods:}
While both ERM and VREx generally identify the correct region of interest in successful cases, neither method is immune to attention drift. We observe instances in both training paradigms where the model focuses on irrelevant areas, such as healthy fibroglandular tissue or background artifacts (e.g., rows 5 and 6 in Figure \ref{fig:1_resnet34d_good_cases}). This indicates that while VREx is designed to penalize instability, it does not guarantee perfect anatomical focus compared to the unconstrained ERM baseline.


\section{Conclusion}
We conducted a comprehensive evaluation of domain-invariant learning for whole-mammogram classification. Despite the theoretical appeal of methods like Invariant Risk Minimization (IRM) and Variance Risk Extrapolation (VREx), our extensive experiments on public datasets reveal their limitations in practice. IRM proved difficult to stabilize, and VREx failed to outperform the Standard ERM baseline. Our results suggest that simple aggregation of diverse training data via ERM remains the most effective strategy for this task. This study serves as a rigorous benchmark, indicating that future progress in breast cancer prediction requires novel generalization techniques capable of outperforming this strong standard baseline in out-of-domain settings.



\vspace{12pt}
\bibliographystyle{IEEEtran}
\bibliography{paper}

@article{shen2019deep,
  title={Deep learning to improve breast cancer detection on screening mammography},
  author={Shen, Li and others},
  journal={Scientific reports},
  volume={9},
  number={1},
  pages={1--12},
  year={2019},
  publisher={Nature Publishing Group}
}

@article{lee2017curated,
  title={A curated mammography data set for use in computer-aided detection and diagnosis research},
  author={Lee, Rebecca Sawyer and others},
  journal={Scientific data},
  volume={4},
  number={1},
  pages={1--9},
  year={2017},
  publisher={Nature Publishing Group}
}

@article{jeong2023emory,
  title={The EMory BrEast imaging Dataset (EMBED): A racially diverse, granular dataset of 3.4 million screening and diagnostic mammographic images},
  author={Jeong, Jiwoong J and Vey, Brianna L and Bhimireddy, Ananth and Kim, Thomas and Santos, Thiago and Correa, Ramon and Dutt, Raman and Mosunjac, Marina and Oprea-Ilies, Gabriela and Smith, Geoffrey and others},
  journal={Radiology: Artificial Intelligence},
  volume={5},
  number={1},
  pages={e220047},
  year={2023},
  publisher={Radiological Society of North America}
}

@article{khaled2022categorized,
  title={Categorized contrast enhanced mammography dataset for diagnostic and artificial intelligence research},
  author={Khaled, Rana and Helal, Maha and Alfarghaly, Omar and Mokhtar, Omnia and Elkorany, Abeer and El Kassas, Hebatalla and Fahmy, Aly},
  journal={Scientific Data},
  volume={9},
  number={1},
  pages={122},
  year={2022},
  publisher={Nature Publishing Group UK London}
}

@article{moreira2012inbreast,
  title={Inbreast: toward a full-field digital mammographic database},
  author={Moreira, In{\^e}s C and others},
  journal={Academic radiology},
  volume={19},
  number={2},
  pages={236--248},
  year={2012},
  publisher={Elsevier}
}

@article{loizidou2021digital,
  title={Digital subtraction of temporally sequential mammograms for improved detection and classification of microcalcifications},
  author={Loizidou, Kosmia and Skouroumouni, Galateia and Pitris, Costas and Nikolaou, Christos},
  journal={European radiology experimental},
  volume={5},
  number={1},
  pages={1--12},
  year={2021},
  publisher={SpringerOpen}
}

@inproceedings{moura2013benchmarking,
  title={Benchmarking datasets for breast cancer computer-aided diagnosis (CADx)},
  author={Moura, Daniel Cardoso and others},
  booktitle={Iberoamerican Congress on Pattern Recognition},
  pages={326--333},
  year={2013},
  organization={Springer}
}

@misc{Strand2022CSAW-CC,
	doi={10.5878/45vm-t798},
	language={en},
	publisher={Karolinska Institutet},
	title={{CSAW-CC (mammography)}},
	url={https://doi.org/10.5878/45vm-t798},
	author={Strand, Fredrik},
	date=2022,
	year=2022,
}

@inproceedings{sechopoulos2021artificial,
  title={Artificial intelligence for breast cancer detection in mammography and digital breast tomosynthesis: State of the art},
  author={Sechopoulos, Ioannis and Teuwen, Jonas and Mann, Ritse},
  booktitle={Seminars in Cancer Biology},
  volume={72},
  pages={214--225},
  year={2021},
  organization={Elsevier}
}

@article{sharma2021retrospective,
  title={Retrospective large-scale evaluation of an AI system as an independent reader for double reading in breast cancer screening},
  author={Sharma, Nisha and Ng, Annie Y and James, Jonathan J and Khara, Galvin and Ambrozay, Eva and Austin, Christopher C and Forrai, Gabor and Fox, Georgia and Glocker, Ben and Heindl, Andreas and others},
  journal={medRxiv},
  pages={2021--02},
  year={2021},
  publisher={Cold Spring Harbor Laboratory Press}
}

@inproceedings{he2016deep,
  title={Deep residual learning for image recognition},
  author={He, Kaiming and Zhang, Xiangyu and Ren, Shaoqing and Sun, Jian},
  booktitle={Proceedings of the IEEE conference on computer vision and pattern recognition},
  pages={770--778},
  year={2016}
}

@inproceedings{liu2022convnet,
  title={A convnet for the 2020s},
  author={Liu, Zhuang and Mao, Hanzi and Wu, Chao-Yuan and Feichtenhofer, Christoph and Darrell, Trevor and Xie, Saining},
  booktitle={Proceedings of the IEEE/CVF Conference on Computer Vision and Pattern Recognition},
  pages={11976--11986},
  year={2022}
}

@article{dosovitskiy2020image,
  title={An image is worth 16x16 words: Transformers for image recognition at scale},
  author={Dosovitskiy, Alexey and Beyer, Lucas and Kolesnikov, Alexander and Weissenborn, Dirk and Zhai, Xiaohua and Unterthiner, Thomas and Dehghani, Mostafa and Minderer, Matthias and Heigold, Georg and Gelly, Sylvain and others},
  journal={arXiv preprint arXiv:2010.11929},
  year={2020}
}

@article{arjovsky2019invariant,
  title={Invariant risk minimization},
  author={Arjovsky, Martin and Bottou, L{\'e}on and Gulrajani, Ishaan and Lopez-Paz, David},
  journal={arXiv preprint arXiv:1907.02893},
  year={2019}
}

@inproceedings{krueger2021out,
  title={Out-of-distribution generalization via risk extrapolation (rex)},
  author={Krueger, David and Caballero, Ethan and Jacobsen, Joern-Henrik and Zhang, Amy and Binas, Jonathan and Zhang, Dinghuai and Le Priol, Remi and Courville, Aaron},
  booktitle={International Conference on Machine Learning},
  pages={5815--5826},
  year={2021},
  organization={PMLR}
}

@article{gulrajani2020search,
  title={In search of lost domain generalization},
  author={Gulrajani, Ishaan and Lopez-Paz, David},
  journal={arXiv preprint arXiv:2007.01434},
  year={2020}
}

@article{vapnik1991principles,
  title={Principles of risk minimization for learning theory},
  author={Vapnik, Vladimir},
  journal={Advances in neural information processing systems},
  volume={4},
  year={1991}
}

@article{wu2019deep,
  title={Deep neural networks improve radiologists’ performance in breast cancer screening},
  author={Wu, Nan and Phang, Jason and Park, Jungkyu and Shen, Yiqiu and Huang, Zhe and Zorin, Masha and Jastrzebski, Stanislaw and Fevry, Thibault and Katsnelson, Joe and Kim, Eric and others},
  journal={IEEE transactions on medical imaging},
  volume={39},
  number={4},
  pages={1184--1194},
  year={2019},
  publisher={IEEE}
}

@article{barros2022virtual,
  title={Virtual Biopsy by Using Artificial Intelligence--based Multimodal Modeling of Binational Mammography Data},
  author={Barros, Vesna and Tlusty, Tal and Barkan, Ella and Hexter, Efrat and Gruen, David and Guindy, Michal and Rosen-Zvi, Michal},
  journal={Radiology},
  pages={220027},
  year={2022},
  publisher={Radiological Society of North America}
}

@article{de2023impact,
  title={Impact of Different Mammography Systems on Artificial Intelligence Performance in Breast Cancer Screening},
  author={de Vries, Clarisse F and Colosimo, Samantha J and Staff, Roger T and Dymiter, Jaroslaw A and Yearsley, Joseph and Dinneen, Deirdre and Boyle, Moragh and Harrison, David J and Anderson, Lesley A and Lip, Gerald and others},
  journal={Radiology: Artificial Intelligence},
  volume={5},
  number={3},
  pages={e220146},
  year={2023},
  publisher={Radiological Society of North America}
}

@article{yala2021toward,
  title={Toward robust mammography-based models for breast cancer risk},
  author={Yala, Adam and Mikhael, Peter G and Strand, Fredrik and Lin, Gigin and Smith, Kevin and Wan, Yung-Liang and Lamb, Leslie and Hughes, Kevin and Lehman, Constance and Barzilay, Regina},
  journal={Science Translational Medicine},
  volume={13},
  number={578},
  pages={eaba4373},
  year={2021},
  publisher={American Association for the Advancement of Science}
}

@article{lotter2021robust,
  title={Robust breast cancer detection in mammography and digital breast tomosynthesis using an annotation-efficient deep learning approach},
  author={Lotter, William and Diab, Abdul Rahman and Haslam, Bryan and Kim, Jiye G and Grisot, Giorgia and Wu, Eric and Wu, Kevin and Onieva, Jorge Onieva and Boyer, Yun and Boxerman, Jerrold L and others},
  journal={Nature Medicine},
  volume={27},
  number={2},
  pages={244--249},
  year={2021},
  publisher={Nature Publishing Group US New York}
}

@article{wang2022domain,
  title={Domain Invariant Model with Graph Convolutional Network for Mammogram Classification},
  author={Wang, Churan and Li, Jing and Sun, Xinwei and Zhang, Fandong and Yu, Yizhou and Wang, Yizhou},
  journal={arXiv preprint arXiv:2204.09954},
  year={2022}
}
\end{document}